\documentclass{article}
\usepackage{spconf,amsmath,graphicx}

\usepackage{multirow}
\title{Almost-unsupervised Speech Recognition with Close-to-zero Resource Based on Phonetic Structures Learned from Very Small Unpaired Speech and Text Data}
\name{Yi-Chen Chen, Chia-Hao Shen, Sung-Feng Huang, Hung-yi Lee, Lin-shan Lee}
\address{National Taiwan University, Taiwan}
\begin{document}
\ninept
\maketitle
%

%%%%%%%%% ABSTRACT
\begin{abstract}
Producing a large amount of annotated speech data for training ASR systems remains difficult for more than 95\% of languages all over the world which are low-resourced.
However, we note human babies start to learn the language by the sounds of a small number of exemplar words without hearing a large amount of data.
We initiate some preliminary work in this direction in this paper.
Audio Word2Vec is used to obtain embeddings of spoken words which carry phonetic information extracted from the signals.
An autoencoder is used to generate embeddings of text words based on the articulatory features for the phoneme sequences.
Both sets of embeddings for spoken and text words describe similar phonetic structures among words in their respective latent spaces.
A mapping relation from the audio embeddings to text embeddings actually gives the word-level ASR.
This can be learned by aligning a small number of spoken words and the corresponding text words in the embedding spaces.
In the initial experiments only 200 annotated spoken words and one hour of speech data without annotation gave a word accuracy of 27.5\%, which is low but a good starting point.
\end{abstract}
%

%%%%%%%%% INTRODUCTION
\section{Introduction}
\label{sec:intro}

Huge success in automatic speech recognition (ASR) has been achieved and widely used in many daily applications.
However, with the existing technologies, machines have to learn from a large amount of annotated data to achieve acceptable accuracy, which makes the development of such technology for a new language with very low resource challenging. 
Collecting a large amount of speech data is expensive, not to mention having the data annotated.
This remains true for at least 95\% of languages all over the world.
Although substantial efforts have been reported on semi-supervised ASR~\cite{karita2018semi,vesely2013semi,dikici2016semi,thomas2013deep,grezl2014combination,vesely2017semi}, in most cases a large amount of speech data including a good portion annotated were still needed.
So training ASR systems with very small data, most of which are not annotated, is an important but unsolved problem.

On the other hand, we note human babies start to learn the language by the sounds of a small number of exemplar words without hearing a large amount of data.
It is certainly highly desired if machines can do that too.
In this paper we initiate some preliminary work in this direction.

Audio Word2Vec has been proposed recently to transform spoken words (signal segments for words without knowing the underlying word it represents) to vectors of fixed dimensionality~\cite{chung2016audio}.
These vectors are shown to carry the information about the phonetic structures of the spoken words.
Segmental Audio Word2Vec was further proposed to jointly segment an utterance into a sequence of spoken words and transform them into a sequence of vectors~\cite{SSAE}.
Unsupervised phoneme recognition was then achieved to some extent by making the duration of spoken words shorter and mapping the clusters of vector representations to phonemes even without aligned audio and text~\cite{liu2018completely}.

To further achieve word-based speech recognition in a very low resource setting, algorithms similar to the skip-gram  model~\cite{mikolov2013distributed} were used to capture the contextual information in speech data and obtain a set of audio semantic embeddings~\cite{chen2018towards,chen2018phonetic,chung2018unsupervised}.
A transformation matrix is then learned to align the audio semantic embeddings with the semantic embeddings learned from text, based on which speech recognition can be performed.
In this way, two semantic embedding spaces are aligned, and speech recognition is performed.
However, semantic information has to be learned from the context out of extremely large amount of data, such as hundreds of hours of speech data~\cite{chen2018towards,chung2018unsupervised}, while rarely used words still have relatively poor embeddings for lack of contextual information.
Achieving ASR through the semantic spaces of speech and text seems to be a too far route for the low-resource requirements considered here.

In this work, we discover the phonetic structures among text words and spoken words in their respective embedding spaces with only a small amount of text and speech data without annotation, and then align the text and speech embedding spaces using a very small number of annotated spoken words.
The word accuracy achieved in the initial experiments with only 200 annotated spoken words and  one hour of audio without annotation is 27.5\%, which is still relatively low but a good starting point.

\begin{figure}[t]
  \centering
  \includegraphics[width=\linewidth]{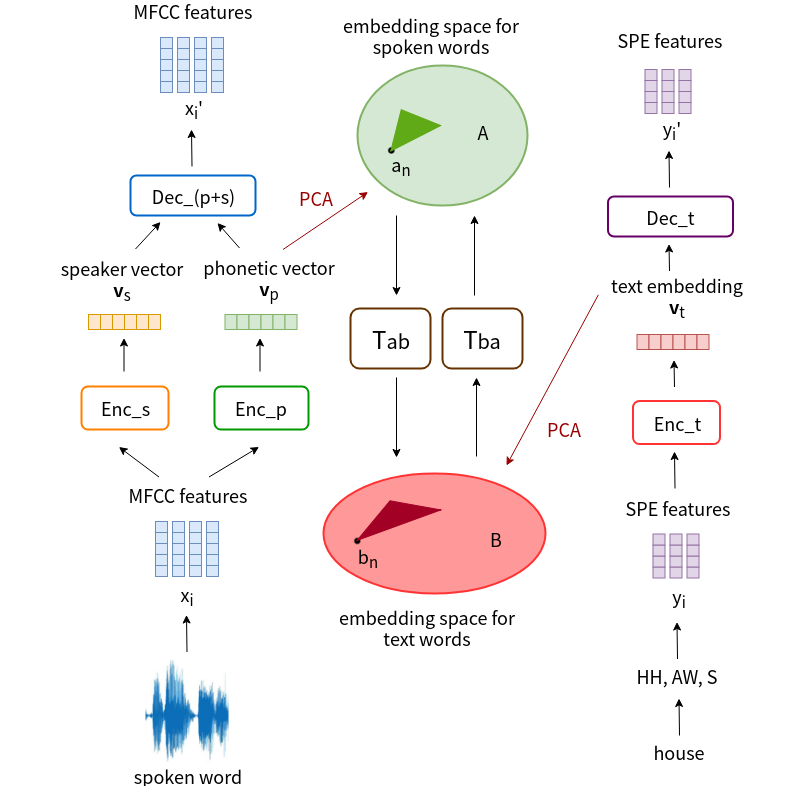}
  \caption{The whole architecture of the proposed approach.}
  \label{fig:overview}
\end{figure}

\section{Proposed Approaches}
In Sections~\ref{sec:a_pho_emb} and~\ref{sec:t_pho_emb}, phonetic embeddings of spoken and text words are respectively learned from audio and text data with relatively small size.
The two embedding spaces are then aligned with a very small set of annotated spoken words in Section~\ref{sec:align}, so speech recognition can be achieved.
In Section~\ref{sec:rescore}, a very weak language model is used to rescore the recognition results.
 
%%%%%%%%% PHONETIC EMBEDDING OF SPOKEN WORDS
\subsection{Phonetic Embedding of Spoken Words}
\label{sec:a_pho_emb}

We denote the speech corpus as $\mathbf{X} = {\{\mathbf{x}_{i}\}}_{i=1}^{M}$, which consists of $M$ spoken words, each represented as $\mathbf{x}_i=(\mathbf{x}_{i_1}, \mathbf{x}_{i_2}, ..., \mathbf{x}_{i_T})$, where $\mathbf{x}_{i_t}$ is the acoustic feature vector at time t and $T$ is the length of the spoken word. 
In the initial work here we focus on the alignment of words in speech and text, so we assume all training spoken words have been properly segmented with good boundaries.
There exist many approaches which can segment utterances automatically~\cite{SRAILICASSP15,WordEmbedIS14,QbyELSTMICASSP15,settle2017query,SemanticRepresentationICASSP18,scharenborg2010unsupervised}, including the Segmental Audio Word2Vec~\cite{SSAE} mentioned above.

A text word corresponds to many different spoken words with varying acoustic factors such as speaker characteristics, microphone characteristics, background noise.
We jointly refer to them as speaker characteristics here for simplicity. 
We perform phonetic embedding over the spoken words with speaker characteristics disentangled using the recently proposed approach~\cite{chen2018phonetic} as shown in the left part of Fig.~\ref{fig:overview}.

A sequence of acoustic features $\mathbf{x}_i=(\mathbf{x}_{i_1}, \mathbf{x}_{i_2}, ..., \mathbf{x}_{i_T})$ is entered to a phonetic encoder $Enc\_p$ and a speaker encoder $Enc\_s$ to obtain a phonetic vector $\mathbf{v_p}$ and a speaker vector $\mathbf{v_s}$. 
Then the phonetic and speaker vectors $\mathbf{v_p}$, $\mathbf{v_s}$ are used by the decoder $Dec\_(p+s)$ to reconstruct the acoustic features $\mathbf{x}'_i$. 
The two encoders $Enc\_p$, $Enc\_s$ and the decoder $Dec\_(p+s)$ are jointly learned to minimize the mean square error between $\mathbf{x}'_i$ and $\mathbf{x}_i$.

Besides, assuming that the spoken word $\mathbf{x}_i$ is uttered by speaker $s_i$.
When the speaker information is not available, we can simply assume that the spoken words in the same utterance are produced by the same speaker. 
If $\mathbf{x}_i$ and $\mathbf{x}_j$ are uttered by the same speaker ($s_i = s_j$), the speaker encoder $Enc\_s$ should make their speaker vectors $\mathbf{v_s}_i$ and $\mathbf{v_s}_j$ as close as possible.
But if $s_i \neq s_j$, the distance between $\mathbf{v_s}_i$ and $\mathbf{v_s}_i$ should exceed a threshold $\lambda$.

Finally, a speaker discriminator $Dis\_s$ takes two phonetic vectors $\mathbf{v_p}_i$ and $\mathbf{v_p}_j$ as input and tries to tell if the two vectors come from the same speaker. 
An additional learning target of the phonetic encoder $Enc\_p$ is to "fool" this speaker discriminator $Dis\_s$, keeping it from discriminating the speaker identity correctly. 
In this way, only the phonetic information is learned in the phonetic vector $\mathbf{v_p}$, while only the speaker characteristics is encoded in the speaker vector $\mathbf{v_s}$. 
The phonetic vector $\mathbf{v_p}$ will be used in the following alignment process in Section~\ref{sec:align}.

%%%%%%%%% PHONETIC EMBEDDING OF TEXT WORDS
\subsection{Phonetic Embedding of Text Words}
\label{sec:t_pho_emb}

To model the phonetic structures of text words, we represent each text word with a phoneme sequence, and each phoneme represented as a vector of articulatory features~\cite{brondsted1998spe}.
So the articulatory features of a text word is denoted as $\mathbf{y}_i=(\mathbf{y}_{i_1}, \mathbf{y}_{i_2}, ..., \mathbf{y}_{i_L})$, where $\mathbf{y}_{i_l}$ is the vector for the $l$\textsuperscript{th} phoneme and $L$ is the number of phonemes in the word. 
Then we feed $\mathbf{y}_i$ into an autoencoder for text words ($Enc\_t$ and $Dec\_t$) to obtain a text embedding $\mathbf{v_t}$, as in the right part of Fig.~\ref{fig:overview}.
The objective of $Enc\_t$ and $Dec\_t$ is to reconstruct the original articulatory  feature representation.

More specifically, we use 15-dim SPE (The Sound Pattern of English~\cite{brondsted1998spe}) articulatory features, respectively corresponding to the features sonorant, syllabic, consonantal, high, back, front, low, round, tense, anterior, coronal, voice, continuant, nasal, and strident.
A value 1 is assigned to a dimension if a phoneme is positive for the corresponding feature, and a value -1 is assigned if negative.
A value 0 is assigned if a phoneme does not have the corresponding feature.
For example, a word ``house" is represented with a phoneme sequence (HH, AW, S), and the phoneme ``S" can be represented with (-1, -1, 1, -1, 0, 0, 0, 0, 0, 1, 1, -1, 1, -1, 1).

%%%%%%%%% LINEAR ALIGNMENT OF SPEECH AND TEXT
\subsection{Alignment of Speech and Text}
\label{sec:align}

This approach was inspired from the Mini-Batch Cycle Iterative Closest Point (MBC-ICP)~\cite{hoshen2018iterative}. 
Since both $Enc\_p$ and $Enc\_t$ encode the phonetic structures of the words in their respective latent spaces for $\mathbf{v_p}$ and $\mathbf{v_t}$, we may assume in these latent spaces the relative relationships among words should have similar structures (the phonetic structures).
For example, the relationships among the words ``other", ``bother" and ``brother" should be very similar in the two spaces.
Similarly for all other words.
It is therefore possible to align the two spaces with a little annotation.

All the vector representations $\mathbf{v_p}$ and $\mathbf{v_t}$ are denoted as $\mathbf{V_p} = {\{\mathbf{v}_{\mathbf{p}_i}\}}_{i=1}^{M}$ and $\mathbf{V_t} = {\{\mathbf{v}_{\mathbf{t}_i}\}}_{i=1}^{M^\prime}$ respectively, where $M$ is the number of spoken words in the speech corpus, and $M^\prime$ the number of distinct words in the text corpus. These vectors are first normalized to zero mean and unit variance in all dimensions, and then projected onto their lower dimensional space by PCA. 
The projected vectors in the principal component spaces are respectively denoted as $\mathbf{A} = {\{\mathbf{a}_i\}}_{i=1}^{M}$ and $\mathbf{B} = {\{\mathbf{b}_i\}}_{i=1}^{M^\prime}$.
This is shown in the middle part of Fig.~\ref{fig:overview}.
Assuming the relative relationships among vectors remain almost unchanged to a good extent after PCA, PCA makes the following alignment easier.

To align the latent spaces for speech and text, we need a small number of ``seeds" or paired speech and text words, referred to as the training data here, is denoted as $\mathbf{A^\prime} = {\{\mathbf{a}_n\}}_{n=1}^{N}$ and $\mathbf{B^\prime} = {\{\mathbf{b}_n\}}_{n=1}^{N}$, in which $\mathbf{a}_n$ and $\mathbf{b}_n$ correspond to the same word. 
$N$ is a small number, $N \ll M$.
Then a pair of transformation matrices, $\mathbf{T_{ab}}$ and $\mathbf{T_{ba}}$ are learned, where $\mathbf{T_{ab}}$ transforms a vector $\mathbf{a}$ in $\mathbf{A}$ to the space of $\mathbf{B}$, that is, $\tilde{\mathbf{b}} = \mathbf{T_{ab}}\mathbf{a}$, while $\mathbf{T_{ba}}$ maps a vector $\mathbf{b}$ in $\mathbf{B}$ to the space of $\mathbf{A}$.
$\mathbf{T_{ab}}$ and $\mathbf{T_{ba}}$ are initialized as identity matrices and then learned iteratively with gradient descent minimizing the objective function:
\begin{equation}
\begin{aligned}
L = & \sum_{n=1}^N \|\mathbf{b}_n - \mathbf{T_{ab}}\mathbf{a}_n \|_2^2 +  \sum_{n=1}^N \|\mathbf{a}_n - \mathbf{T_{ba}}\mathbf{b}_n \|_2^2  \\
&+ \lambda^\prime \sum_{n=1}^N \| \mathbf{a}_n - \mathbf{T_{ba}}\mathbf{T_{ab}}\mathbf{a}_n \|_2^2 \\
&+ \lambda^\prime \sum_{n=1}^N \| \mathbf{b}_n - \mathbf{T_{ab}}\mathbf{T_{ba}}\mathbf{b}_n \|_2^2 .
\end{aligned} \label{eq:align}
\end{equation} 
In the first two terms, we want the transformation of $\mathbf{a}_n$ by $\mathbf{T_{ab}}$ to be close to $\mathbf{b}_n$ and vice versa.
The last two terms are for cycle consistency, i.e., after transforming $\mathbf{a}_n$ to the space of $\mathbf{B}$ by $\mathbf{T_{ab}}$ and then transforming back by $\mathbf{T_{ba}}$, it should end up with the original $\mathbf{a}_n$, and vice versa.
$\lambda^\prime$ is a hyper-parameter. 

After we obtain the transformation matrix $\mathbf{T_{ab}}$, it can be directly applied on all phonetic vectors $\mathbf{v_p}$.
For each spoken word with phonetic vector $\mathbf{v_p}_i$, we can get its PCA mapped vector $\mathbf{a}_i$ and the transformed vector $\mathbf{b}_i = \mathbf{T_{ab}}\mathbf{a}_i$.
Then we can find the nearest neighbor vector of $\mathbf{b}_i$ among all vectors in $\mathbf{B}$, whose corresponding text word is the result of alignment, or ASR result.

%%%%%%%%% RESCORING WITH A LANGUAGE MODEL
\subsection{Rescoring with a Language Model} 
\label{sec:rescore}

\begin{figure}[t]
  \centering
  \includegraphics[width=\linewidth]{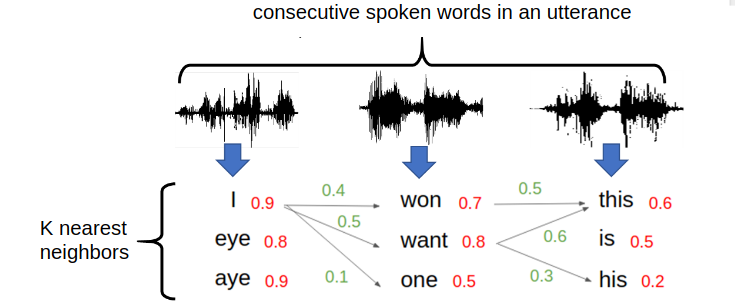}
  \caption{Language model rescoring with beam search for top $3$ words ($K=3$).
  The numbers in red are cosine similarities for embeddings, while those in green are from the language model.}
  \label{fig:beam_search}
\end{figure}

We can use a language model to perform rescoring as illustrated in Figure~\ref{fig:beam_search}.
During testing, for each spoken word $\mathbf{a}_i$, we obtain a vector $\mathbf{b}_i = \mathbf{T_{ab}}\mathbf{a}_i$.
We find the $K$ nearest neighbor vectors of $\mathbf{b}_i$, $\{\mathbf{b}_k\}_{k=1}^K$, among the vectors in $\mathbf{B}$. 
Each $\mathbf{b}_k$ corresponds to a text word, with the cosine similarity between $\mathbf{b}_i$ and $\mathbf{b}_k$ (numbers in red in Figure~\ref{fig:beam_search}) as the score. 
The language model scores are the numbers in green.
Any kind of language modeling can be used here, and the cosine similarities and language model scores are integrated to find the best path in beam search.

\begin{figure}[t]
  \centering
  \includegraphics[width=\linewidth]{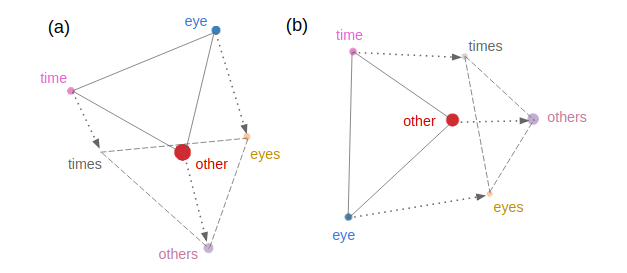}
  \caption{3-dim visualization for the first three principal component vectors of several words in the embedding spaces of (a) $\mathbf{v_p}$ and (b) $\mathbf{v_t}$.}
  \label{fig:PCA}
\end{figure}

\begin{table}[t]
%\footnotesize
\small
\centering
\caption{Top-1 and top-10 accuracies with different N (numbers of paired words).}
\label{table:pair_num}
\begin{tabular}{|c|c|c|c|c|}
\hline

\multicolumn{1}{|c|}{N} &
\multicolumn{2}{c|}{Paired acc. (\%)} &
\multicolumn{2}{c|}{Unpaired acc. (\%)} \\ \hline

\multicolumn{1}{|c|}{no. of pairs} &
\multicolumn{1}{c|}{top-1} &
\multicolumn{1}{c|}{top-10} &
\multicolumn{1}{c|}{top-1} &
\multicolumn{1}{c|}{top-10} \\ \hline \hline

1533 (all) & 71.3 & 95.2 & 15.5 & 34.7 \\ \hline
1000 & 82.4 & 98.7 & 14.9 & 33.1 \\ \hline
500 & 92.7 & 99.7 & 14.5 & 31.3 \\ \hline
300 & 98.6 & 100.0 & 15.8 & 34.0 \\ \hline
200 & 100.0 & 100.0 & \bf{17.2} & \bf{35.5} \\ \hline
100 & 100.0 & 100.0 & 14.8 & 28.9 \\ \hline
50 & 100.0 & 100.0 & 8.1 & 19.4 \\ \hline
20 & 100.0 & 100.0 & 2.4 & 8.4 \\ \hline

\end{tabular}
\end{table}

\begin{table}[t]
%\footnotesize
\small
\centering
\caption{
%\textcolor{blue}{
The results of beam search with different beam widths.
%}
}
\label{table:beam_search}
%{\color{blue}
\begin{tabular}{|c|c|c|}
\hline
\multicolumn{1}{|c|}{Beam width} &
%\multicolumn{1}{c|}{End-to-end (\%)} &
\multicolumn{1}{c|}{Top-1 acc. (\%)}\\ \hline \hline

no language model & 17.2 \\ \hline
1 & 21.5 \\ \hline
3 & 24.9 \\ \hline
10 & 26.4 \\ \hline
50 & 27.5 \\ \hline

\end{tabular}
%}
\end{table}

\begin{table}[t]
%\footnotesize
\small
\centering
\caption{Examples of beam search with beam width K=50.}
\label{table:BS_examples}
\begin{tabular}{|c|c|c|}
\hline

\multirow{3}{*}{(1)} &
before & ... \textbf{ned it feet mather} flowers \textbf{out} all ... \\ \cline{2-3}
& after & ... and if she gathered flowers \textbf{out} all ... \\ \cline{2-3}
& reference & ... and if she gathered flowers at all ... \\ \hline \hline

\multirow{3}{*}{(2)} &
before & ... \textbf{cur} mother \textbf{whiff shrill} in \textbf{zee} her ... \\ \cline{2-3} %\textbf{aunty two whitney} 
& after & ... her mother was three in \textbf{the} her ... \\ \cline{2-3} %only \textbf{two guinea} 
& reference & ... her mother was three in teaching her ... \\ \hline \hline %only to repeat 

\multirow{3}{*}{(3)} &
before & ... \textbf{key beganne} to \textbf{pearl} her hair ... \\ \cline{2-3} %\textbf{nabbed} long for \textbf{piles} 
& after & ... she began to \textbf{tell} her \textbf{head} ... \\ \cline{2-3} %and \textbf{longed} for \textbf{while} 
& reference & ... she began to curl her hair ...\\ \hline %and long for balls 

\end{tabular}
\end{table}

\begin{table}[t]
%\footnotesize
\small
\centering
\caption{Ablation studies for disentanglement of speaker characteristics and SPE representation for text words.}
\label{table:ablation}
\begin{tabular}{|c|c|c|c|c|c|}
\hline

\multicolumn{1}{|c|}{Disent.} &
\multicolumn{1}{c|}{Text} &
\multicolumn{2}{c|}{Paired acc.} &
\multicolumn{2}{c|}{Unpaired acc.} \\ \hline

\multicolumn{1}{|c|}{} &
\multicolumn{1}{c|}{} &
\multicolumn{1}{c|}{top-1} &
\multicolumn{1}{c|}{top-10} &
\multicolumn{1}{c|}{top-1} &
\multicolumn{1}{c|}{top-10} \\ \hline \hline

yes & SPE & 100.0 & 100.0 & \bf{17.2} & \bf{35.5} \\ \hline
no & SPE & 100.0 & 100.0 & 15.2 & 31.3 \\ \hline
yes & one-hot & 100.0 & 100.0 & 17.0 & 35.5 \\ \hline
no & one-hot & 100.0 & 100.0 & 15.0 & 32.1 \\ \hline

\end{tabular}
\end{table}

%%%%%%%%% EXPERIMENTS
\section{Experiments}
\label{sec:exp}

%%%% DATASET
\subsection{Dataset}
\label{subsec:dataset}

%Lee: 我這樣講對嗎？
LibriSpeech~\cite{panayotov2015librispeech} for read speech in English derived from audiobooks were used here. 
It contains 1000 hours of speech uttered by 2484 speakers.
We used one-hour speech from the ``clean-100" set as the training data for phonetic embedding of spoken words. 
This training set contains a total of 9022 spoken words for 1533 distinct words.
In the initial experiments here we used word boundaries from LibriSpeech, although joint learning of segmentation and spoken word embedding is achievable~\cite{SSAE}.
39-dim MFCCs were taken as the acoustic features.  
The ``clean-100" set of 100 hours of speech includes 32219 distinct words.
We trained phonetic embedding of text words on all these 32219 words.
For the ``seeds" or paired data, we selected the $N$ most frequent text words out of the 9022 in the one-hour training data, and randomly chose a corresponding spoken word from the one-hour set for each text word to form the paired training set, $\mathbf{A^\prime}$ and $\mathbf{B^\prime}$.　
In this way, each word had at most one example in the paired data.
The alignment or ASR results were evaluated on the whole one-hour data set, but excluding those used in the paired data.

%%%% MODEL IMPLEMENTATION
\subsection{Model Implementation}
\label{subsec:implementation}

The phonetic encoder $Enc\_p$, speaker encoder $Enc\_s$ and text encoder $Enc\_t$ were all single-layer Bi-GRUs with hidden layer size 256.
The decoders $Dec\_(p+s)$ and $Dec\_t$ were double-layer GRUs with hidden layer size 512 and 256 respectively.
The speaker discriminator $Dis\_s$ was a fully-connected feedforward network with 2 hidden layers of size 256.
The value of threshold $\lambda$ for the constraint of $\mathbf{v_s}$ in Section~\ref{sec:a_pho_emb} was set to 0.01.
$\lambda^\prime$ in Equation (\ref{eq:align}) of Section~\ref{sec:align} was set to 0.5.
In all training processes, the initial learning rate was 0.0001 and the mini-batch size was 64.

%%%% VISUALIZATION OF PCA VECTORS
\subsection{Visualization of PCA Vectors}
\label{subsec:PCA}

We visualized several typical examples of vectors $\mathbf{v_p}$ and $\mathbf{v_t}$ in the subspaces of the first three PCA components.
For cleaner visual presentation, we only show the averages of the vectors corresponding to all spoken words for the same text word.
In Fig.~\ref{fig:PCA} (a) and (b) respectively for $\mathbf{v_p}$ and $\mathbf{v_t}$, we show the six vectors corresponding to the six words \{time, eye, other, times, eyes, others\}.
We see the triangle structures for the three words and the difference vectors between those for ``word" and ``word-s" are quite consistent in both spaces.
Many similar examples can be found, including those for ``verb" and ``verb-ing", etc.
These results illustrate the phonetic structures among spoken and text words offer a good knowledge source for ASR, and PCA provides very good initialization for such processes.

%%%% RESULTS
\subsection{Results}
\label{subsec:results}

In the following tables, we use top-1 and top-10 accuracies as the evaluation metrics for the alignment or word-level ASR.
The columns under ``Paired acc." are the accuracies for the training paired spoken and text words, and those under ``Unpaired acc." are the achieved recognition accuracies of testing spoken words. 

\subsubsection{The Effect of N (Number of Paired Words) Without Language Model}
\label{subsubsec:pair_num}

The results for different numbers of paired words $N$ without language model are listed in Table~\ref{table:pair_num}.
We see the paired accuracy was not 100\% for N $>$ 200.
For example, it was 82.4\% for N = 1000.
This implies the linear mapping matrix $\mathbf{T_{ab}}$ was not adequate for transforming 1000 pairs even if the transformation was already given.
Better transformation is needed and currently under progress.
This explained why training with N = 200 up to 1533 paired words didn't offer higher unpaired accuracies.
In other words, 200 paired words already provide all information the present framework of a linear transformation matrix $\mathbf{T_{ab}}$ can learn.
The highest top-1 unpaired accuracy of 17.2\% for N = 200 seemed low since 200 / 1533 = 13.0\% paired words were given.
Note here each text word may correspond to many spoken words, but only one of them were given as the training pair.
We'll also see the accuracy can be improved further with a language model.
Also, when N $\leq$ 50, the accuracies drop dramatically obviously because too few clues were available in learning the mapping relations.
All the following experiments were for N = 200.

\subsubsection{Rescoring with a Language Model}
\label{subsubsec:BS}

We used the transcriptions of the ``clean-100" set (excluding the one-hour data set) to train a very weak bi-gram language model. 
The beam width for beam search ranged from 1 to 50.
We simply summed the cosine similarities and the language model scores with weights 1 and 0.05 during the rescoring.
The results 
%\textcolor{blue}{
%of our model using beam search
%} 
are listed in 
%\textcolor{blue}{
%the third column of
%} 
Table~\ref{table:beam_search}.
We see the very weak language model offered very good improvements (27.5\% vs 17.2\% for K = 50), verifying the mapping from spoken to text words performed the word-level recognition to a certain extent.
Some examples showing how the beam search helped (beam width K = 50) in Table~\ref{table:BS_examples}.
We see the language model may correct some phonetically similar but semantically different words.
It is believed with a stronger language model and a well-designed fusion approach, the accuracy can be further improved.

%%
%\subsubsection{Comparison with  Semi-supervised ASR} %an End-to-end
%\label{subsubsec:compare}

%\textcolor{blue}{
%To compare the proposed model with other semi-supervised approaches, we trained a sequence-to-sequence model~\cite{sutskever2014sequence} taking a spoken word as input and predicting a phoneme sequence.
%The training data was the 200 annotated spoken words and one hour of annotated speech data, which is the same data setting used in the training of our model.
%The self-training~\cite{grandvalet2005semi} method was used here.
%The model was first trained on the annotated data and then used to generate pseudo labels for the unannotated data.
%Then the pseudo labels were used to fine-tune the model.
%}

%\textcolor{blue}{
%The results of this model using beam search are listed in the second column of Table~\ref{table:beam_search}.
%Because a sophisticated mapping must be learned to map input features to words with the  model, severe overfitting occurred due to little paired data even with self-training.
%However, our model unsupervisedly learned meaningful phonetic embedding representations of speech and text with autoencoders respectively, and aligned them with a simple linear transformation, which had no excessive complexity and avoided overfitting.
%}

%% 
\subsubsection{Ablation Studies}
\label{subsubsec:ablation}

Here we further verify that the disentanglement of speaker characteristics for $\mathbf{v_p}$ and the SPE-based representation used in text word embedding both improved the performance in Table~\ref{table:ablation} (N = 200 without language model).
Removing the disentanglement and/or representing the phonemes with one-hot vectors instead of SPE-based features led to some performance degradation.

%%%%%%%%% CONCLUSION
\section{Conclusion}
\label{sec:end}
In this work, we let the machines learn the phonetic structures among spoken words and text words based on their embeddings, and learn the mapping from spoken words to text words based on the transformation between the two phonetic structures using a small number of annotated words.
Initial experiments with only one hour of speech data and 200 annotated spoken words gave an accuracy of 27.5\%.
There is still a long way to go, and a very wide space to be explored.

\bibliographystyle{IEEEbib}
\bibliography{mybib,IR_bib,ref_dis,segment,transfer,INTERSPEECH16,ICASSP13,refs}

\end{document}